\documentclass{article}
\usepackage{cite}
\usepackage{amsmath,amssymb,amsfonts}
\usepackage{algorithmic}
\usepackage{graphicx}
\usepackage{algorithm,algorithmic}
\usepackage{hyperref}
\hypersetup{hidelinks=true}
\usepackage{textcomp}

\usepackage{arxiv}

\usepackage[utf8]{inputenc} 
\usepackage[T1]{fontenc}    
\usepackage{hyperref}       
\usepackage{url}            
\usepackage{booktabs}       
\usepackage{amsfonts}       
\usepackage{nicefrac}       
\usepackage{microtype}      
\usepackage{cleveref}       
\usepackage{lipsum}         
\usepackage{graphicx}
\usepackage[numbers]{natbib}
\usepackage{doi}

\usepackage{tabularx}
\usepackage{siunitx}
\usepackage{multirow}
\usepackage{colortbl}

\usepackage{array}
\usepackage{makecell}

\newcolumntype{P}[1]{>{\centering\arraybackslash}p{#1}}

\usepackage[]{mdframed}
\definecolor{gray75}{gray}{.25}
\definecolor{gray70}{gray}{.3}
\definecolor{gray60}{gray}{.4}
\definecolor{gray50}{gray}{.5}
\definecolor{gray40}{gray}{.6}
\definecolor{gray30}{gray}{.7}
\definecolor{gray20}{gray}{.8}
\definecolor{gray15}{gray}{.85}
\definecolor{gray10}{gray}{.9}
\definecolor{gray05}{gray}{.95}
\definecolor{steBoxLine}{rgb}{0.0, 0.0, 0.0}
\mdfdefinestyle{stebox1}{%
linecolor=steBoxLine,
linewidth=3pt,%
leftmargin=0cm,rightmargin=0cm,%
roundcorner=5pt,
topline=false,bottomline=false,rightline=false,%
backgroundcolor=gray05
}

\mdfdefinestyle{prompt}{%
    linecolor=steBoxLine,
    linewidth=3pt,%
    leftmargin=0cm,rightmargin=0cm,%
    roundcorner=5pt,
    topline=false,bottomline=false,rightline=false,%
    backgroundcolor=gray05, skipabove=.5em, skipbelow=.5em, font=\sffamily\small  
}

\newenvironment{prompt}[1]
  {\begin{mdframed}[style=prompt,frametitle={\sffamily\small\textbf{#1}}]}
  {\end{mdframed}}

\title{HELIOT: LLM-Based CDSS for Adverse Drug Reaction Management}

\author{ \href{https://orcid.org/0000-0002-1153-1566}{\includegraphics[scale=0.06]{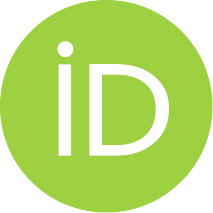}\hspace{1mm}De Vito Gabriele}\\
        Software Engineering (SeSa) Lab\\
        University of Salerno\\
        Salerno, Italy\\
	\texttt{gadevito@unisa.it} \\
	\And
        \href{https://orcid.org/0000-0002-0975-8972}{\includegraphics[scale=0.06]{orcid.pdf}\hspace{1mm}Ferrucci Filomena}\\
        Software Engineering (SeSa) Lab\\
        University of Salerno\\
        Salerno, Italy\\
	\texttt{fferrucci@unisa.it} \\
	\And
	\href{https://orcid.org/00000-0003-1226-9560}{\includegraphics[scale=0.06]{orcid.pdf}\hspace{1mm}Athanasios Angelakis} \\
	Department of Epidemiology and Data Science\\
	Amsterdam UMC Locatie AMC\\
	Amsterdam, The Netherlands \\
        Digital Health; Methodology;\\
        Amsterdam Public Health Research Institute\\ 
        Amsterdam, The Netherlands \\
        University of Amsterdam\\
        Data Science Center\\
        Amsterdam, The Netherlands\\  
	\texttt{a.angelakis@amsterdamumc.nl}
}

\renewcommand{\headeright}{}
\renewcommand{\undertitle}{}
\renewcommand{\shorttitle}{HELIOT}

\hypersetup{
pdftitle={Design and Evaluation of a CDSS for Drug Allergy Management Using LLMs and Pharmaceutical Data Integration},
pdfauthor={De Vito G., Ferrucci F., Angelakis A.},
pdfkeywords={Clinical Decision Support Systems, Large Language Models, Drug Administration, Allergy Management },
}

\begin{document}

\maketitle

\begin{abstract}
Medication errors significantly threaten patient safety, leading to adverse drug events and substantial economic burdens on healthcare systems. Clinical Decision Support Systems (CDSSs) aimed at mitigating these errors often face limitations when processing unstructured clinical data, including reliance on static databases and rule-based algorithms, frequently generating excessive alerts that lead to alert fatigue among healthcare providers. This paper introduces HELIOT, an innovative CDSS for adverse drug reaction management that processes free-text clinical information using Large Language Models (LLMs) integrated with a comprehensive pharmaceutical data repository. HELIOT leverages advanced natural language processing capabilities to interpret medical narratives, extract relevant drug reaction information from unstructured clinical notes, and learn from past patient-specific medication tolerances to reduce false alerts, enabling more nuanced and contextual adverse drug event warnings across primary care, specialist consultations, and hospital settings. 
An initial evaluation using a synthetic dataset of clinical narratives and expert-verified ground truth shows promising results. HELIOT achieves high accuracy in a controlled setting. 
In addition, by intelligently analyzing previous medication tolerance documented in clinical notes and distinguishing between cases requiring different alert types, HELIOT can potentially reduce interruptive alerts by over 50\% compared to traditional CDSSs. 
While these preliminary findings are encouraging, real-world validation will be essential to confirm these benefits in clinical practice. 
\end{abstract}

\section{Introduction}
\label{sec:intro}
Medication errors pose significant risks to patient safety and lead to adverse drug events (ADEs) \cite{elliott2021economic}. In England, an estimated 237 million such errors occur annually, with around 66 million being potentially clinically significant \cite{elliott2021economic, ahsani2022interventions}. These incidents cost the National Health System (NHS) approximately £98.5 million annually, consume 181,626 bed days, and contribute to 1,708 deaths \cite{elliott2021economic}. Similarly, in the United States, the economic impact is substantial, with prescribing and administration mistakes 
costing an estimated \$20 billion annually 
\cite{ahsani2022interventions}.

CDSSs have emerged as tools to mitigate medication errors and enhancing patient safety \cite{sutton2020overview}. These systems provide healthcare professionals with evidence-based recommendations and alerts, helping to prevent potential ADEs. However, traditional CDSSs 
face several limitations \cite{kwan2020computerised, sarkar2020effective, meunier2023barriers}. They typically rely on static databases and rule-based algorithms, which may not capture the nuances of individual patient cases or the latest medical knowledge \cite{sarkar2020effective, meunier2023barriers, quan2023usefulness}. 
For instance, when a drug with potential cross-reactions is prescribed, these systems generate alerts without considering complex clinical scenarios documented in notes, such as previous patient-specific tolerance, distinctions between true allergies and minor side effects, or situation-specific risk-benefit analyses \cite{colicchio2023beyond}.  
Evidence shows that incorporating this contextual information could significantly reduce unnecessary alerts  \cite{colicchio2023beyond}. 
Without such capabilities, the rigid alerting mechanisms lead to excessive warnings, contributing to alert fatigue among healthcare providers and potentially causing critical warnings to be overlooked \cite{ meunier2023barriers, quan2023usefulness, colicchio2023beyond, van2021optimizing, westerbeek2021barriers, van2022overall, topaz2016rising, topaz2015high, hsieh2004characteristics}.

The advent of Large Language Models (LLMs), such as GPT-4 \cite{achiam2023gpt}, offers a promising avenue to address these limitations. LLMs possess advanced natural language processing (NLP) capabilities, 
and have already succeeded in various healthcare applications, such as predicting drug interactions and patient outcomes, assisting in diagnostic processes, and generating clinical notes \cite{tripathi2024efficient, rios2024evaluation, de2025llms}. We conjecture that their ability to process and synthesize large volumes of unstructured data makes them an innovative substitute for current rule-based CDSSs, which often lack the flexibility and depth of understanding required for intricate patient-specific situations. 

This paper presents HELIOT, a novel CDSS for adverse drug reaction management that processes unstructured clinical narratives, building upon our previous exploration of LLM integration in healthcare technologies \cite{de2024assessing}. The system overcomes the shortcomings of traditional CDSSs by leveraging LLMs to interpret 
medical narratives from clinical notes, text-based electronic health records, and transcribed patient conversations. 
By integrating this capability with a comprehensive pharmaceutical data repository, 
HELIOT improves the accuracy and reliability of adverse reaction alerts in healthcare. 
Our preliminary evaluation suggests that HELIOT shows promise in providing accurate 
recommendations while 
reducing alert fatigue.

The primary contributions of this paper are as follows.
\begin{itemize}
\setlength{\itemindent}{-5pt}
    \item We present a novel approach to process unstructured clinical narratives for adverse drug reaction management, demonstrating how LLMs can extract and interpret patient-specific medication reaction information, including medication tolerances, adverse events, and cross-sensitivities.
    \item We describe the HELIOT CDSS's modular architecture, highlighting its core components and the approach employed to provide decision support services. 
    \item We present an empirical evaluation of the HELIOT CDSS, comparing its performance with a ground truth provided by medical experts.
    \item We provide datasets and code used to develop the HELIOT prototype, and the synthetic patient dataset employed for the empirical evaluation as a contribution to the research community.
\end{itemize}
\smallskip
\noindent \textbf{Structure of the paper.} The remainder of this paper is organized as follows: Section \ref{sec:bkg_related} describes the background on CDSSs and LLMs and the related work. Section \ref{sec:methods} describes our method, including the data pipeline, the HELIOT 
approach, and the empirical study design used to evaluate the proposed CDSS. Section \ref{sec:results} shows and discusses the empirical results. Section \ref{sec:discussion} provides the practical implications,  future research lines, and limitations of our study. Section \ref{sec:conclusions} concludes the paper.

\section{State of Art and Motivation}
\label{sec:bkg_related}
This section reviews current CDSS challenges and limitations, discusses recent advances in LLMs for processing medical text, and presents the motivation behind our work.

\subsection{CDSSs Challenges and Limitations}
CDSSs are pivotal in improving patient safety and clinical efficiency through integration with Electronic Medical Records \cite{sutton2020overview}. In the domain of pharmacology, these systems must navigate particularly complex territory, managing intricate drug interactions and patient-specific factors that require sophisticated algorithms to detect potential adverse reactions \cite{roy2022drug}. Despite their potential benefits, their implementation in the pharmacological field faces several challenges \cite{sutton2020overview}.

Research demonstrates the value of CDSSs in reducing medication errors, with knowledge-based systems showing a 4.4\% improvement in prescribing behavior \cite{kwan2020computerised} and AI-based approaches achieving enhanced accuracy in prescription verification \cite{corny2020machine}. However, a persistent challenge emerges specifically with drug reactions alert systems, where override rates remain problematically high—ranging from 43.7\% to 97\% \cite{luri2022systematic}, with particularly concerning rates for opioids \cite{topaz2015high}. This widespread dismissal of alerts has been consistently documented across multiple studies \cite{colicchio2023beyond, meunier2023barriers, quan2023usefulness, van2021optimizing, westerbeek2021barriers, van2022overall, topaz2016rising, topaz2015high, hsieh2004characteristics}, suggesting a fundamental limitation in current approaches. 

Promising solutions have emerged, including ontology-based systems \cite{calvo2022ontopharma} and machine learning approaches \cite{rozenblum2020using}, but these methods fall short when representing complex clinical scenarios or achieving satisfactory alert acceptance rates. Evidence indicates that incorporating previous drug tolerance data could significantly reduce unnecessary alerts \cite{colicchio2023beyond}, highlighting a critical gap: current systems cannot effectively interpret unstructured clinical narratives that contain crucial patient-specific information about medication tolerances and reactions. This constraint in processing natural language documentation represents an opportunity where advanced natural language processing techniques could potentially transform alert relevance and clinical acceptance.

\subsection{LLMs Opportunities}
LLMs have revolutionized the field of NLP by leveraging the Transformer architecture with self-attention mechanisms, as introduced by Vaswani et al.~\cite{vaswani_2017}. Notable examples of these models include commercial offerings such as GPT-3~\cite{gpt3_2020} and GPT-4~\cite{achiam2023gpt}, as well as open-source models like BERT~\cite{bert_2018}, FLAN-T5~\cite{chung2022scaling}, LLama~\cite{touvron2023llama}, BLOOM~\cite{scao_2022_bloom}, and GLM~\cite{zeng_2022_glm}. LLMs are trained on extensive text datasets and often contain hundreds of billions of parameters~\cite{chen2021evaluating, KASNECI2023102274, zhao2023survey}. The initial "pre-training" phase is computationally intensive but essential for enabling these models to perform a wide range of NLP tasks, such as translation and summarization, with high proficiency~\cite{chen2021evaluating, KASNECI2023102274, zhao2023survey}. 

Following pre-training, LLMs can be specialized through a fine-tuning process, which involves using smaller datasets to tailor the models for specific NLP tasks, such as question-answering or tasks within different domains. Several emergent abilities have been discovered in the context of LLMs. The key abilities include "In-context learning," "Instruction following," and "Step-by-step reasoning." "In-context learning," introduced by GPT-3, allows models to perform tasks based on examples without additional training. "Instruction following" enables the models to execute tasks based solely on given instructions, while "Step-by-step reasoning" facilitates solving complex problems through chain-of-thought prompting~\cite{shanahan2023talking}. 
These sophisticated capabilities make LLMs particularly promising for healthcare applications, with recent research demonstrating their effectiveness in generating differential diagnoses \cite{rios2024evaluation} and enhancing clinical documentation \cite{tripathi2024efficient}. Their unique ability to process context-rich information and perform complex reasoning aligns with the challenge of interpreting medication narratives in clinical notes—a critical area where current CDSSs fail. Despite this natural fit, applying LLMs to extract and interpret medication information from unstructured clinical text remains underexplored.

\subsection{Motivation of our Work}
The potential of LLMs for healthcare applications is clear, but translating this potential into practical systems for medication safety requires addressing several domain-specific challenges. 
Even in healthcare settings with sophisticated Electronic Health Record systems (EHR), clinical notes documenting patient reactions to drugs, referral letters detailing treatment histories, and records of medication tolerances frequently remain as free-text narratives. Traditional CDSSs can only process structured data and cannot interpret these narrative texts \cite{colicchio2023beyond}. This means that valuable information about a patient's drug history may be overlooked when prescribing medications.  
For instance, a clinical note stating "Patient experienced mild rash with amoxicillin but has since tolerated cephalexin" contains important information about drug allergies and tolerances, but traditional CDSSs cannot extract and use this information because it is in free-text format. LLMs may help bridge this gap by providing the contextual understanding needed to interpret such narrative texts.

In addition, the research community lacks comprehensive evaluation datasets 
for assessing CDSSs in this field  \cite{van2022overall}. Our work aims to fill these gaps by providing open-source tools and evaluation resources to support future research in enhancing clinical decision support through natural language understanding.

\section{HELIOT Framework}
\label{sec:methods}
This section presents the HELIOT CDSS, illustrating  design principles and implementation details.

\subsection{CDSS Design and Approach}
\label{sec:cdss}
In the following, we detail the decision support process, system architecture, and adaptability features that form the basis of our CDSS framework.

\subsubsection{Decision Support Process}
HELIOT employs a Retrieval Augmentation Generation (RAG) approach to support physicians in medication decisions based on patients' adverse reaction histories. RAG is a technique that enhances LLMs by retrieving relevant information from external knowledge sources before generating responses, making the output more informed, contextualized, and accurate.
The decision process follows several integrated steps to produce clinical assessments.

The foundation of our approach begins with the parallel retrieval of drug information from specialized databases containing pharmaceutical data (active ingredients, excipients, contraindications, and side effects) while simultaneously analyzing patient clinical notes to identify potentially problematic ingredients. 

\begin{figure}[h]
\begin{prompt}{Decision Support Prompt: System Prompt}\fontsize{3mm}{4mm}\selectfont
Act as an expert physician.\\
Your task is to check if the drug I want to prescribe is safe for the patient, focusing only on the potential drug reactions  the patient has.\\
\#\#\# Drug To Prescribe: $\{drug\}$\\
\#\#\# Drug Active Ingredients: $\{active\_ingredients\}$\\
\#\#\# Drug Excipients: $\{excipients\}$\\
\#\#\# Known Cross-reactivity: $\{cross\_reactivity\}$\\
\#\#\# Known Excipients With Chemical Cross-reactivity:\\
$\{excipients\_cross\_reacts\}$\\
\#\#\# Contraindications: $\{contraindications\}$\\\\
\#\# INSTRUCTIONS \#\#\\
...\\
\#\# OUTPUT FORMAT \#\#\\
\{"a":"brief description of your analysis", "r":"final response: NO DOCUMENTED REACTIONS OR INTOLERANCES|DIRECT ACTIVE INGREDIENT REACTIVITY|DIRECT EXCIPIENT REACTIVITY|NO REACTIVITY TO PRESCRIBED DRUG'S INGREDIENTS OR EXCIPIENTS|CHEMICAL-BASED CROSS-REACTIVITY TO EXCIPIENTS|DRUG CLASS CROSS-REACTIVITY WITHOUT DOCUMENTED TOLERANCE|DRUG CLASS CROSS-REACTIVITY WITH DOCUMENTED TOLERANCE", "rt":"reaction type: None|Life-threatening|Non life-threatening immune-mediated|Non life-threatening non immune-mediated"\}
\end{prompt}
\centering
\caption{Decision Support Prompt: System Prompt.}
\label{fig:sysinputbase}
\end{figure}

\begin{figure}[h!]
\begin{prompt}{Decision Support Prompt: User Prompt}\fontsize{3mm}{4mm}\selectfont
\#\#\# PATIENT INFORMATION: $\{clinical\_notes\}$
\end{prompt}
\centering
\caption{Decision Support Prompt: User Prompt.}
\label{fig:usrinputbase}
\end{figure}

A critical feature of our system is its ability to maintain continuity of care by integrating current and historical patient data. The patient database stores and updates clinical notes from previous encounters, creating a longitudinal record of adverse reactions and tolerances. When a healthcare professional consults the system about a new medication, these historical records are automatically retrieved and combined with current clinical notes. This integration provides a complete picture of the patient's reaction history, even in facilities without integrated EHR systems, ensuring that past adverse events are not overlooked in current decision-making. Once all relevant information, including drug composition, current clinical notes, and the patient's historical data, is gathered, 
the system executes the core decision support logic. This logic employs carefully crafted prompts (Figures \ref{fig:sysinputbase} and \ref{fig:usrinputbase}) that guide the LLM to analyze the clinical situation. Following the persona pattern \cite{white2023prompt}, our system instructs the LLM to embody an expert physician who evaluates potential adverse reactions by examining relationships between the drug's composition and the patient's documented reaction history, including current and historical notes.

The system's final output provides an assessment structured in three parts: a clinical classification of the case (e.g., "Direct Active Ingredient Reactivity"), a categorization of the reaction severity (e.g., "Life-threatening"), and a detailed analysis explaining the rationale behind these classifications. This structured approach ensures that physicians receive clear, actionable guidance based on patient data while understanding the underlying clinical reasoning.

\begin{figure}[h]
\begin{prompt}{Ingredient Translation Prompt}
\fontsize{3mm}{4mm}\selectfont
Translate in English from {language}: $\{text\}$
Report only the translation, nothing else. If you don't know the translation, report the original text.
\end{prompt}
\centering
\caption{Ingredient Translation Prompt.}
\label{fig:translation}
\end{figure}

It is worth noting that HELIOT faces a significant linguistic challenge when processing clinical notes from different regions and healthcare systems. Current LLMs, as demonstrated by Wendler et al \cite{wendler2024llamas}, exhibit an intrinsic bias toward English in their conceptual processing, potentially compromising reasoning accuracy when working with non-English medical terminology. HELIOT addresses this limitation by standardizing all medical and pharmaceutical terminology into English using a dedicated translation prompt (Figure \ref{fig:translation}). This approach not only overcomes the language bias of underlying models but also ensures optimal correspondence with international medical ontologies that predominantly use English nomenclature.

\subsubsection{System Architecture}
\label{sec:cdssarch}
\begin{figure}[h]
    \centering
    \includegraphics[width=0.8\linewidth]{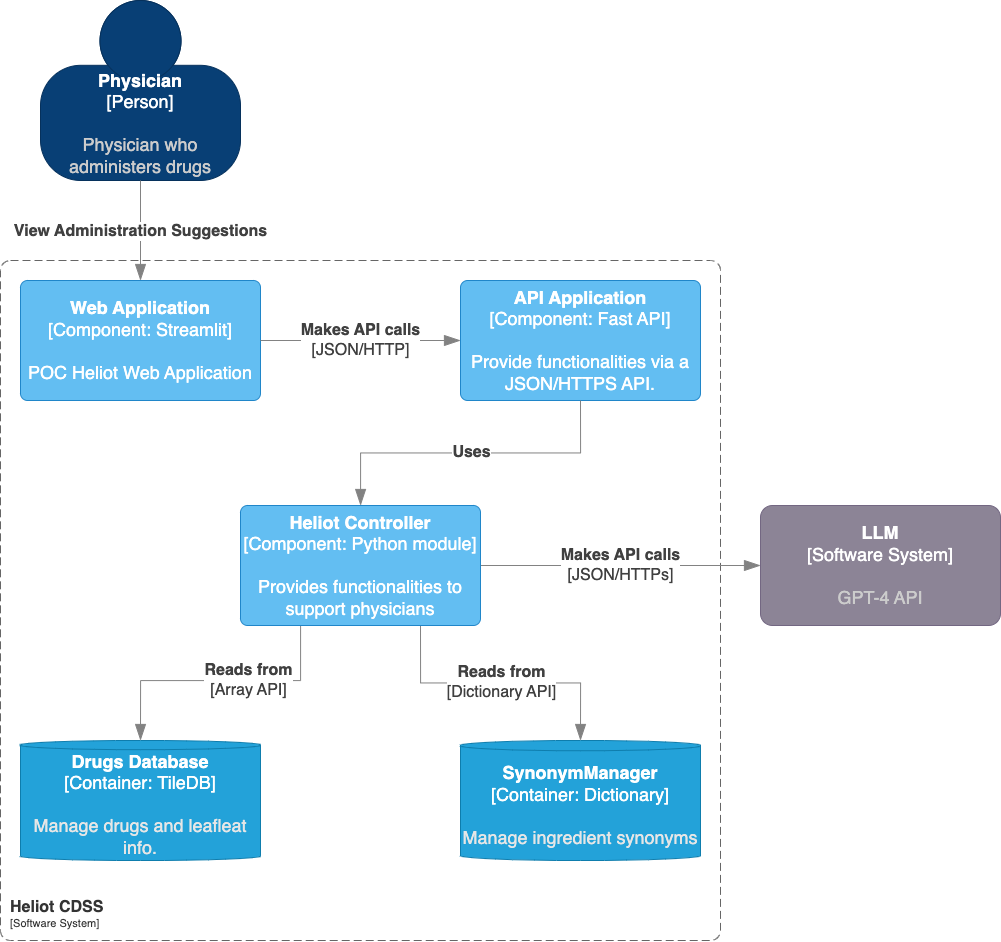}
    \caption{HELIOT CDSS Architecture.}
    \label{fig:cdss_arc}
\end{figure}

Building upon the approach described above, we designed HELIOT with an integrated, modular architecture that translates our methodology into a practical, deployable system. This architecture functions as both a standalone solution and a service that can be integrated with existing EHR systems.  
The system's core components, which operationalize the RAG approach and historical data integration described earlier, include a web application, an API application, and a central controller, as shown in Figure \ref{fig:cdss_arc}.

The web application serves as the user interface for standalone operation, while the API application provides RESTful services that can be consumed either by our web interface or by existing EHR systems through standard integration protocols. This service-oriented architecture enables HELIOT to function as an external decision support service that commercial EHRs can invoke without replacing their native CDSS infrastructure. 

The API application streams responses in real-time, similar to ChatGPT, reducing perceived latency by providing immediate feedback as results become available. 

The HELIOT Controller is the core decision-making engine. It uses several sub-components to retrieve and process data, such as the TileDB Drug and Patient Databases, the in-memory Synonym Manager, and the LLM.

The database component stores comprehensive drug-related information and patient clinical notes, while an in-memory structure manages ingredient synonyms. 
More specifically, the pharmaceutical data required to populate the drug database includes drug identifiers, active ingredients, excipients, contraindications, and side effects. Except for drug identifiers, this information does not need to follow a rigid structure and can be stored as free text. However, for ingredient management, HELIOT requires a standardized approach where each canonical ingredient name is associated with its corresponding list of synonyms.
We selected TileDB \cite{papadopoulos2016tiledb}, which offers efficient data retrieval, storage compression capabilities, and versatility in supporting both local and cloud deployments, ensuring scalability across different environments. 

The LLM is employed for advanced natural language processing tasks, such as interpreting 
medical texts and generating contextually relevant recommendations.
This modular architecture ensures that each component can be independently updated or scaled, providing flexibility and robustness.

\subsubsection{Data Flexibility and Adaptability}
HELIOT is designed with inherent flexibility in both its approach and architecture. The pharmaceutical database accommodates data from diverse sources without requiring extensive reformatting. As detailed in section \ref{sec:cdssarch}, while drug identifiers need standardization, most critical information—including active ingredients, excipients, contraindications, and side effects—can be stored as unstructured free text, eliminating the need for complex data normalization procedures. 
Similarly, the patient clinical database—while maintained locally within HELIOT—demonstrates adaptability in processing clinical information. The system does not impose rigid requirements on structuring clinical notes, relying instead on the LLM's natural language processing capabilities to extract relevant adverse reaction information from various documentation styles and formats.   
Regarding linguistic adaptation, the standardization process employs generalized language conversion mechanisms to normalize terminology into a standard language. This approach allows healthcare facilities in regions with different primary languages to utilize the system while maintaining consistent clinical reasoning. 

Finally, the database technology selected for HELIOT offers flexibility through support for different deployment models (local or cloud-based) and efficient query mechanisms that maintain performance even as the database grows. 
All this means that HELIOT can serve as a CDSS in various healthcare contexts, from settings with advanced commercial EHRs to those with limited technological infrastructure, while maintaining its core architecture and local databases.

\subsection{HELIOT Prototype}
\label{sec:cdsspoc}
This section describes our implementation of the HELIOT CDSS design as a functional prototype. 

\subsubsection{Pharmaceutical Data Pipeline}
\label{sec:data}
\begin{figure}[h]
    \centering
    \includegraphics[width=\linewidth]{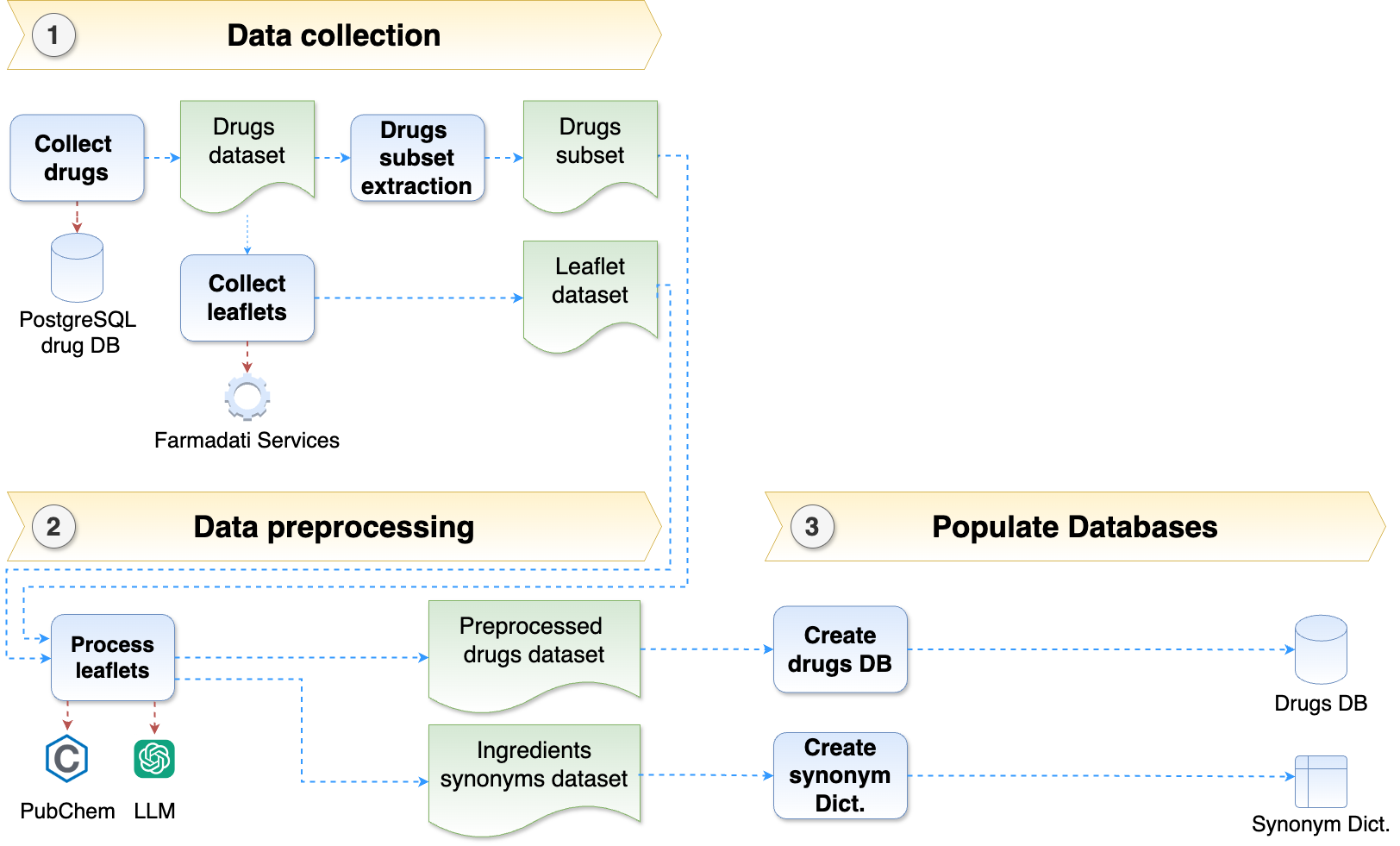}
    \caption{Overview of the Pharmaceutical Data Pipeline Process.}
    \label{fig:data_pipeline}
\end{figure}
The first step in developing our prototype was creating a comprehensive pharmaceutical knowledge base. 
To this aim, we developed a specialized data pipeline to transform raw pharmaceutical data into HELIOT's pharmaceutical knowledge base. 
The pipeline (see Figure \ref{fig:data_pipeline}) consists of three main phases: 1) Data Collection, 2) Data Preprocessing, and 3) Database Population.

The subsequent subsections offer an explanation of each phase. 

\paragraph{Data collection}
In the data collection phase, we gathered the necessary data for our experiments, specifically the datasets for drugs and leaflets.

\subparagraph{Drug dataset}
The first step was creating the drug dataset. To this end, an Italian company provided us with the PostgreSQL dump of the database of Italian medicines approved by the Italian Medicines Agency (AIFA) \footnote{AIFA. \url{https://www.aifa.gov.it/en/trova-farmaco}}. 
The drug dataset has 106,962 drugs. 
Table \ref{table:drug_dataset} reports the dataset columns.

\begin{table} [h]
    \scriptsize
    \centering 
    \caption{Drug Dataset Structure.}
    \vspace{1mm} 
    \resizebox{1\linewidth}{!}{    
    \begin{tabular}{p{0.2\linewidth} p{0.8\linewidth}}
     
    \rowcolor{black}
    \textcolor{white}{Column} & \textcolor{white}{Description} \\
    \hline
Drug\_code & It is the ministerial code of the drug, also known as the AIC code (Marketing Authorization issued by AIFA)\\
\rowcolor{gray10}
Drug\_name & It is the drug name\\
Drug\_form\_full\_descr & It is the pharmaceutical form of the drug\\
\rowcolor{gray10}
Atc\_code & It is the drug's ATC (Anatomical Therapeutic Chemical) code, according to the Word Health Organization classification system.\\
Leaflet & It is the leaflet file name.\\
\hline

    \end{tabular}
    }
    \label{table:drug_dataset}
\end{table}

\subparagraph{Leaflets dataset}
The next step was to collect the leaflets associated with the medications. To make the pipeline production-ready, we automated this step by referring to a pharmaceutical data provider, the Farmadati company \footnote{Farmadati. \url{https://www.farmadati.it/default.aspx}} that is one of Italy's major pharmaceutical data providers. It offers paid services to download the up-to-date drug database, including leaflets. 
Consequently, we utilized the web services provided by Farmadati to acquire leaflets for each medication in the dataset. The final leaflets dataset contains 19,188 files. 
The leaflets' structure includes the medication's name, qualitative and quantitative composition, pharmaceutical form, clinical information, pharmacological properties, pharmaceutical information, radiation dosimetry data for radiopharmaceuticals, and instructions on extended preparation and quality control for radiopharmaceuticals.

\subparagraph{Drugs subset extraction}
\label{sec:drugs_subset_extraction}
The final step of the data collection phase was creating a representative subset of drugs for 
the subsequent ``data preprocessing'' step. 
The drug dataset was randomly sampled from the drugs dataset using ATC codes. We followed the distribution reported in literature studies \cite{hsieh2004characteristics, topaz2015high, topaz2016rising, colicchio2023beyond} to ensure the dataset reflected real-world drug prescription and adverse reaction  patterns. The sampling process maintained the proportions of different drug classes as observed in clinical settings, with narcotic analgesics representing the largest group (65\%), followed by antibiotics (15\%), NSAID (5\%), diuretics (2\%), antiplatelet agents (2\%), and other medications (11\%). We also included common excipients associated with both immediate and delayed hypersensitivity reactions, with particular attention to preservatives and common allergens (e.g., polyethylene glycol, polysorbates, benzalkonium chloride). 

\begin{table} [h]
    \scriptsize
    \centering 
    \caption{Drug Distribution in Dataset.}
    \vspace{1mm} 
    \resizebox{1\linewidth}{!}{    
    \begin{tabular}{p{0.3\linewidth} p{0.15\linewidth} p{0.15\linewidth}}
    \rowcolor{black}
    \textcolor{white}{Drug Class} & \textcolor{white}{Perc.} & \textcolor{white}{No. of drugs} \\
    \hline
    Opioids & 65\% & 653\\
    \rowcolor{gray10}
    Antibiotics & 15\% & 152\\
    NSAID & 5\% & 47\\
    \rowcolor{gray10}
    Diuretics & 2\% & 24\\
    Antiplatelet agents & 2\% & 16\\
    \rowcolor{gray10}
    Other & 11\% & 108\\
    \hline
    \textbf{TOTAL} & & \textbf{1,000}\\
    \hline
    \end{tabular}
    }
    \label{table:drug_subset_distribution}
\end{table}

Table \ref{table:drug_subset_distribution} reports the final dataset distribution.

\paragraph{Data preprocessing}
\label{sec:data_preproc}
The data preprocessing phase addressed two key challenges in the leaflet data for our drug subset.

First, single leaflets often contain information for multiple pharmaceutical forms of the same medication (e.g., the ORAMORPH leaflet includes details for both syrup and oral solution forms). Since healthcare professionals prescribe specific forms, we must extract form-specific information while excluding irrelevant details. Second, the original ingredients and excipients were in Italian, requiring translation to English (consistently with the linguistic standardization approach discussed in section \ref{sec:cdss}) to enable synonym matching through international services like PubChem \footnote{PubChem substances. \url{https://pubchem.ncbi.nlm.nih.gov/rest/pug/substance/name/{encoded_ingredient_name}/synonyms/JSON}} \footnote{PubChem compounds. \url{https://pubchem.ncbi.nlm.nih.gov/rest/pug/compound/name/{encoded_ingredient_name}/synonyms/JSON}}. This standardization is paramount as ingredients may appear under different names in medical records (e.g., "acetylsalicylic acid" vs "aspirin"). 

To address these challenges, we leveraged GPT-4o with two specialized prompts 
for processing ingredients and leaflet sections. 
For the sake of readability, we did not report the full prompts that can be found in our online repository \cite{devito_github}. 

During this phase, we processed 1,035 unique ingredients and created an ingredient dictionary by querying PubChem REST services for comprehensive synonym lists. The preprocessing resulted in two structured datasets: "leaflet\_info.csv" containing form-specific drug details and "ingredients\_synonyms.csv" with standardized ingredient names and variants. 
Tables \ref{table:leaflet_info} and \ref{table:ingredients_data} detail the structure of these datasets.

\begin{table} [h]
    \scriptsize
    \centering 
    \caption{Processed Leaflet Dataset Structure.}
    \vspace{1mm} 
    \resizebox{1\linewidth}{!}{    
    \begin{tabular}{p{0.28\linewidth} p{0.72\linewidth}}
     
    \rowcolor{black}
    \textcolor{white}{Column} & \textcolor{white}{Description} \\
    \hline
Drug\_code & It is the ministerial code of the drug.\\
\rowcolor{gray10}
Drug\_name & It is the drug name.\\
Drug\_form & It is the pharmaceutical form of the drug.\\
\rowcolor{gray10}
ATC & It is the ATC code of the drug.\\
Composition & It contains the list of the drug’s active ingredients.\\
\rowcolor{gray10}
Excipients & It contains the list of the drug’s inactive ingredients.\\
Contraindications & It contains the contraindications for the drug.\\
\rowcolor{gray10}
Drug\_interactions & It contains the drug interactions.\\
Side effects & It reports the side effects of the drug.\\
\rowcolor{gray10}
Incompatibilities & It reports the incompatibilities for the drug in the case of concurrent pharmaceutical therapies.\\
\hline

    \end{tabular}
    }
    \label{table:leaflet_info}
\end{table}

\begin{table} [h!]
    \scriptsize
    \centering 
    \caption{Ingredient Synonyms Dataset Structure.}
    \vspace{1mm} 
    \resizebox{1\linewidth}{!}{    
    \begin{tabular}{p{0.14\linewidth} p{0.86\linewidth}}
     
    \rowcolor{black}
    \textcolor{white}{Column} & \textcolor{white}{Description} \\
    \hline
Ingredient & It is the ingredient name extracted from the processed leaflets;\\
\rowcolor{gray10}
English\_name & It is the English translation of the ingredient. \\
Synonyms & List of synonyms separated by ‘\#’ returned by the PubChem REST service.\\
\rowcolor{gray10}
Type & It represents the type of the ingredient, namely active or inactive.\\
\hline

    \end{tabular}
    }
    \label{table:ingredients_data}
\end{table}

To ensure the accuracy of the preprocessed data, two healthcare professionals (a clinical pharmacist and a physician) independently validated a random sample of 20\% of the drugs subset. This sample size was determined following established methodology for validation studies \cite{flahault2005sample} and provides an appropriate statistical power for this application. 
The healthcare professionals focused on verifying that GPT-4o correctly extracted information specific to each pharmaceutical form from the official leaflets. They also verified the accuracy of ingredient and excipient translations from Italian to English, ensuring correct matching with PubChem synonyms.  
The validation established that our preprocessing approach, particularly the designed GPT-4o prompts, effectively isolated and extracted form-specific information from the official leaflets while maintaining data integrity. The high inter-rater agreement (Cohen's kappa \cite{cohen1960coefficient} = 0.95) further supported the reliability of our approach, demonstrating "almost perfect" consensus between clinical experts. 

\paragraph{Populate HELIOT Databases}
\label{sec:cdss_db}
We then populated two 
databases using the preprocessed datasets: the drug database using ``leaflet\_info.csv'' and the synonyms database using ``ingredients\_synonyms.csv'' (see section \ref{sec:data_preproc}).  
For the drug database schema, we defined drug\_code, atc\_code, composition, and excipients as dimensions with ASCII data type, while other columns became array attributes. We applied ``ZstdFilter'' compression \cite{collet2018zstandard} at level 3 for text-heavy attributes to optimize storage and query performance. Given the manageable number of ingredients (1,035) for the synonyms database, we implemented an in-memory data structure to ensure fast retrieval of ingredient synonyms during CDSS processing.

\subsubsection{Prototype Implementation}
Building upon the data pipeline, we developed a fully functional proof-of-concept (POC) prototype of the HELIOT CDSS. This implementation demonstrates how the conceptual architecture described in section \ref{sec:cdssarch} translates into practice with specific technology choices and integration patterns.
While the architecture is designed to be model-agnostic and compatible with various LLMs (e.g., LLaMA), for this implementation we utilized GPT-4o through the OpenAI API. 
The prototype consists of the web application developed using the Streamlit framework, the API application developed using the FAST API and Uvicorn frameworks, and the HELIOT Controller developed using TileDB and the OpenAI API.
The web application provides functionalities for processing single prescriptions with streamed results, uploading entire datasets for batch processing, and downloading results. 

All the scripts, results, and source code are provided in our online repository \cite{devito_github}.

\section{Evaluation Methodology}
\label{sec:experiment}
This section describes the methodology for evaluating HELIOT's effectiveness as a clinical decision-support system, including our research objectives, the creation of a specialized evaluation dataset, and the experimental design.

\subsection{Research Goals}
The primary goal of the empirical assessment was to analyze the effectiveness of the proposed CDSS in two key dimensions: accuracy of clinical decision support and reduction of alert fatigue. The purpose was to provide empirical evidence highlighting the benefits and limitations of the HELIOT CDSS, enabling healthcare professionals to be aware of the strengths and weaknesses they would encounter through its use. 

Specifically, the empirical assessment aimed to address the following research question: 

\begin{prompt}{\textbf{RQ:} How effective is HELIOT CDSS in providing support during drug administration while minimizing alert fatigue?}
\end{prompt} 

This research question encompasses the system's ability to correctly identify potential adverse drug reactions and provide contextually appropriate alerts that do not overwhelm clinicians with unnecessary warnings.

\subsection{Synthetic Patient Dataset Creation}
\label{sec:patientsynth}
To comprehensively evaluate HELIOT, we developed a synthetic patient dataset that represents the variety and complexity of adverse drug reaction scenarios encountered in clinical practice. 
Recent studies \cite{kuhnel2024synthetic, giuffre2023harnessing} support this synthetic data approach, demonstrating its potential to replicate real-world analysis results while maintaining privacy and supporting robust evaluation.

To ensure clinical accuracy and relevance, we involved  two additional healthcare professionals (a clinician and a physician) who complemented our earlier pharmaceutical validation team (see section \ref{sec:data_preproc}) with specialized knowledge in patient reactions and medical documentation. We implemented a human-in-the-loop approach, enhancing both the interpretability and reliability of our system by incorporating expert domain knowledge throughout the development process, as demonstrated in similar healthcare Artificial Intelligence implementations where human expertise guides model decisions and validates outputs \cite{lee2022towards, pais2024large, valente2022interpretability}. Under the guidance of these clinical experts, we generated clinical notes for each case, documenting reaction timing, type (life-threatening, non-life-threatening immune-mediated, or non-immune-mediated), symptoms, drug tolerance, and chemical-based cross-reactivity patterns. The dataset structure (see Table \ref{table:synthetic_structure}) was designed to maintain consistency with real-world electronic health records by incorporating standard clinical fields, medication coding systems, and detailed clinical notes that mirror actual documentation practices in healthcare settings.

\begin{table} [h]
 \scriptsize
 \centering
     \caption{Synthetic Dataset Structure.}
 \vspace{1mm}
     \resizebox{1\linewidth}{!}{
     \begin{tabular}{p{0.2\linewidth} p{0.8\linewidth}} \rowcolor{black} \textcolor{white}{Field} & \textcolor{white}{Description} \\
 \hline
 Patient ID & Unique identifier for each case\\
 \rowcolor{gray10}
 Drug Code & Ministerial code of the prescribed drug\\
 Drug Name & Name of the prescribed drug\\
 \rowcolor{gray10}
 Clinical Note & Detailed patient history including adverse reactions\\
 Classification & Case classification category\\
 \rowcolor{gray10}
 Alert Type & Interruptive/Non-interruptive/No alert\\
 Reaction Type & Severity and nature of reaction\\
 \rowcolor{gray10}
 Prescribed ATC & ATC code of the prescribed drug\\
 \hline
 \end{tabular}
 }
 \label{table:synthetic_structure}
 \end{table} 

Based on distribution patterns from literature \cite{hsieh2004characteristics, topaz2015high, topaz2016rising, colicchio2023beyond}, we generated 1,000 synthetic cases covering various scenarios from direct ingredient reactivity to complex cross-reactivity patterns. The healthcare professionals validated the dataset through independent reviews, achieving an inter-rater agreement of 0.91 (Cohen's kappa). Consensus meetings resolved disagreements, particularly regarding alert type assignments in complex cross-reactivity cases.

\begin{figure}[h]
    \centering
    \includegraphics[width=1.0\linewidth]{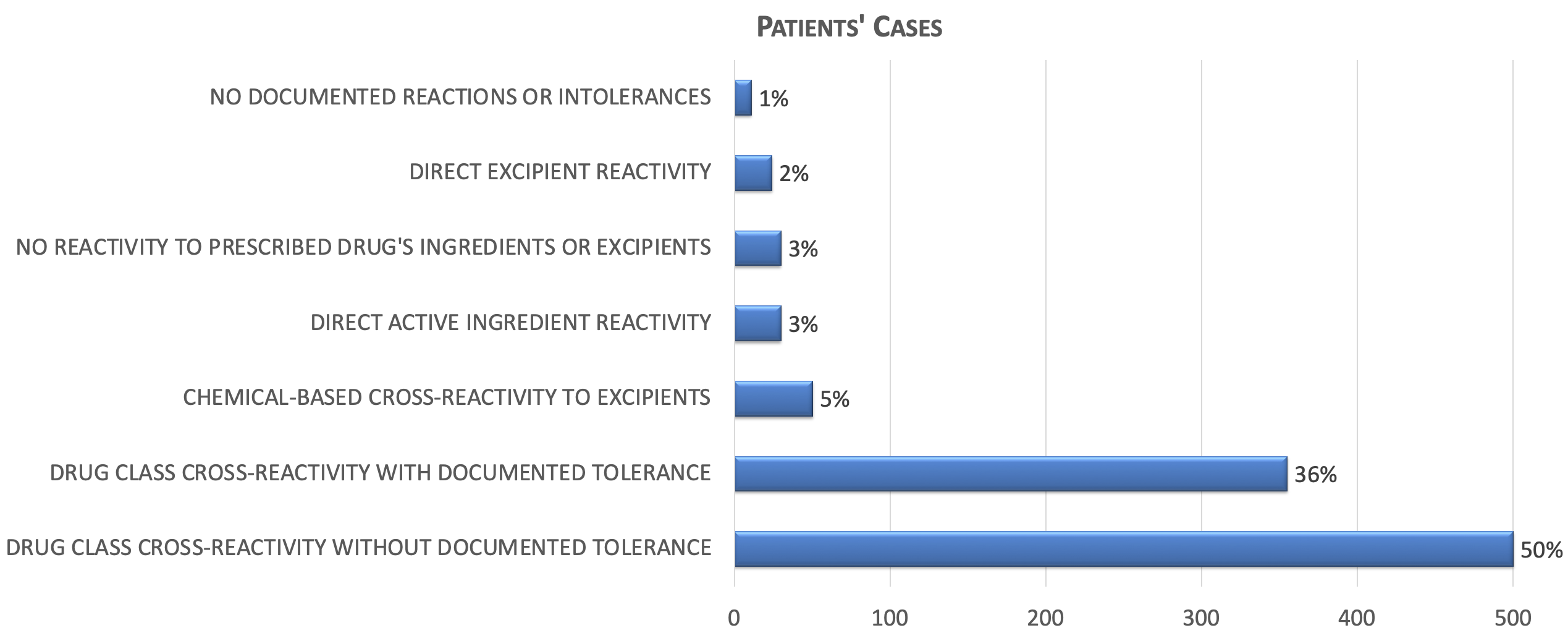}
    \caption{Class Distribution in the Patients Dataset.}
    \label{fig:class_distribution}
\end{figure}

\begin{table} [h]
    \scriptsize
    \centering 
    \caption{Clinical Cases Classifications.}
    \vspace{1mm} 
    \resizebox{1\linewidth}{!}{    
    \begin{tabular}{p{0.30\linewidth} p{0.70\linewidth}}
    \rowcolor{black}
    \textcolor{white}{Classification} & \textcolor{white}{Description} \\
    \hline
    No documented reactions or intolerances & Patients with no history of adverse drug reactions or intolerances in their clinical records\\
    \rowcolor{gray10}
    Direct active ingredient reactivity & Patients with documented adverse reactions to the active ingredient of the prescribed drug\\
    Direct excipient reactivity & Patients with documented reactions to an excipient present in the prescribed drug's formulation\\
    \rowcolor{gray10}
    No reactivity to prescribed drug's ingredients or excipients & Patients with documented adverse drug reactions, but not related to any component of the prescribed drug\\
    Chemical-based cross-reactivity to excipients & Patients with adverse reactions to drugs in the same therapeutic class as the prescribed medication, without documented tolerance\\
    \rowcolor{gray10}
    Drug class cross-reactivity without documented tolerance & Patients with adverse reactions to drugs in the same therapeutic class as the prescribed medication, without documented tolerance\\
    Drug class cross-reactivity with documented tolerance & Patients with adverse reactions to drugs in the same therapeutic class but with documented tolerance to the prescribed medication\\
    \hline
    \end{tabular}
    }
    \label{table:classification_descr}
\end{table}

\begin{table} [h!]
    \scriptsize
    \centering 
    \caption{Case Distribution by Classification, Reaction Type, and Alert Type.}
    \vspace{1mm} 
    \resizebox{1\linewidth}{!}{    
    \begin{tabular}{p{0.35\linewidth} p{0.25\linewidth} p{0.20\linewidth} p{0.05\linewidth} p{0.05\linewidth}}
    \rowcolor{black}
    \textcolor{white}{Case Classification} & \textcolor{white}{Reaction Type} & \textcolor{white}{Alert Type} & \textcolor{white}{Cases} & \textcolor{white}{Perc.} \\
    \hline
    No documented reactions or intolerances & None & None & 11 & 1.1\%\\
    \rowcolor{gray10}
    No reactivity to prescribed drug's ingredients or excipients & None & None & 30 & 3.0\%\\
    \multirow{3}{*}{Direct active ingredient reactivity} & Life-threatening & Interruptive & 9 & 0.9\%\\
    \rowcolor{gray10}
     & Non life-threatening immune-mediated & Interruptive & 12 & 1.2\%\\
     & Non life-threatening non immune-mediated & Non-interruptive & 9 & 0.9\%\\
    \rowcolor{gray10}
    \multirow{3}{*}{Direct excipient reactivity} & Life-threatening & Interruptive & 6 & 0.6\%\\
     & Non life-threatening immune-mediated & Interruptive & 4 & 0.4\%\\
    \rowcolor{gray10}
     & Non life-threatening non immune-mediated & Non-interruptive & 14 & 1.4\%\\
    Chemical-based cross-reactivity to excipients & Life-threatening & Interruptive & 15 & 1.5\%\\
    \rowcolor{gray10}
     & Non life-threatening immune-mediated & Interruptive & 9 & 0.9\%\\
     & Non life-threatening non immune-mediated & Non-interruptive & 26 & 2.6\%\\
    \rowcolor{gray10}
    Drug class cross-reactivity without documented tolerance & Life-threatening & Interruptive & 103 & 10.3\%\\
      & Non life-threatening immune-mediated & Interruptive & 271 & 27.1\%\\
    \rowcolor{gray10}
     & Non life-threatening non immune-mediated & Non-interruptive & 126 & 12.6\%\\
    Drug class cross-reactivity with documented tolerance & None & None & 355 & 35.5\%\\
    \hline
    \textbf{TOTAL} & & & \textbf{1,000} & \textbf{100\%}\\
    \hline
    \end{tabular}
    }
    \label{table:case_distribution_full}
\end{table}

The final dataset includes three alert categories: interruptive alerts (45.5\%), non-interruptive alerts (14.9\%), and no alerts needed (39.6\%). These cases are further categorized into seven clinical classes based on the complexity of adverse drug reactions and sensitivities. 

Figure \ref{fig:class_distribution} shows the distribution of clinical patterns, while Tables \ref{table:classification_descr} and \ref{table:case_distribution_full} detail the classifications and case distribution by reaction and alert types.

\subsection{Experimental Design}
\begin{figure}[h]
    \centering
    \includegraphics[width=1\linewidth]{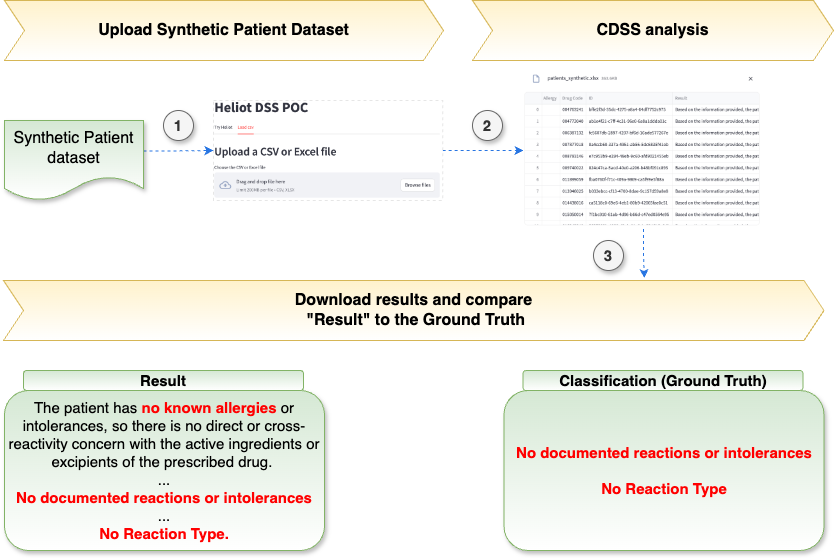}
    \caption{Experiment Design and Setup.}
    \label{fig:cdss_expt}
\end{figure}

To investigate the research question, we compared responses from HELIOT CDSS against ground truth data provided by healthcare professionals in the data pipeline (see Section \ref{sec:patientsynth}). This comparison was made using the HELIOT CDSS prototype web application (see Section \ref{sec:cdsspoc}) for interaction and validation. Figure \ref{fig:cdss_expt} illustrates the design of the experiment.

We first uploaded the synthetic patient dataset (described in Section \ref{sec:patientsynth}) to the HELIOT DSS POC, where the data is analyzed for potential reactions. 
The system's results were then downloaded and compared to the ground truth, evaluating separately the ``Classification'' and ``Reaction Type'' assignments provided by the healthcare professionals. The alert type is automatically determined based on case classification and reaction type, as follows: when there is no documented tolerance, life-threatening reactions and non life-threatening immune-mediated reactions trigger interruptive alerts, while non life-threatening non immune-mediated reactions produce non-interruptive alerts. No alerts are needed when there are no documented reactions or when tolerance has been previously established (see Table \ref{table:case_distribution_full} in section \ref{sec:patientsynth}). 

To assess the consistency of the LLM's responses and evaluate any potential non-deterministic behavior, we repeated the experiment five times under identical conditions, averaged the results, and calculated the Fleiss' Kappa, which is a statistical measure that evaluates the agreement between multiple raters (or iterations, in this case), across iterations \cite{fleiss2013statistical}. This approach allowed us to verify whether the system produces consistent results across multiple runs, which is crucial for ensuring reliability in clinical applications.

To quantitatively measure the effectiveness of the HELIOT CDSS, we employed three key metrics, which are standard in evaluating classification systems and provide a comprehensive view of performance, specifically: Precision, Recall, and F1-score.

Precision is the ratio of true positives (correctly identified instances of a class) to all instances predicted as that class.
Recall measures the proportion of true positive results among all actual positive cases, and indicates how well the system can identify positive instances.
Finally, F1-score is the harmonic mean of precision and recall, providing a metric that balances both concerns. It is particularly useful when the class distribution is imbalanced. 

We also measured the execution time of the analysis along the five runs and averaged the results to evaluate the proposed CDSS's performance comprehensively.
We executed the experiment using a MacBook with a 2 GHz Intel i5 quad-core processor and 16 GB of RAM.

\section{Results}
\label{sec:results}
The results of our evaluation, averaged over five runs to account for potential LLM non-deterministic behavior, demonstrate the high performance of our system. 
The Fleiss kappa score across all iterations showed \textit{perfect} agreement (100\%) \cite{fleiss2013statistical,zampetti2022empirical} for both category and reaction types classifications. 
This perfect agreement confirms that our system exhibited deterministic behavior, producing identical outputs across all five runs despite the potential for LLM-based variability.

\begin{table} [h]
 \scriptsize
 \centering
     \caption{Classification Results per Category (averaged over five runs).} \vspace{1mm}
     \resizebox{1\linewidth}{!}{     \begin{tabular}{p{0.66\linewidth} p{0.07\linewidth} p{0.07\linewidth} p{0.05\linewidth} p{0.05\linewidth}} \rowcolor{black} \textcolor{white}{Case Category} & \textcolor{white}{Precision} & \textcolor{white}{Recall} & \textcolor{white}{F1} & \textcolor{white}{Cases} \\
 \hline
 Chemical-based cross-reactivity to excipients & 0.9804 & 1.0000 & 0.0.9901 & 50\\
 \rowcolor{gray10}
 Direct active ingredient reactivity & 1.0000 & 1.0000 & 1.0000 & 30\\ 
Direct excipient reactivity & 1.0000 & 0.9583 & 0.9787 & 24\\ 
\rowcolor{gray10}
 Drug class cross-reactivity with documented tolerance & 1.0000 & 1.0000 & 1.0000 & 355\\
 Drug class cross-reactivity without documented tolerance & 1.0000 & 1.0000 & 1.0000 & 500\\
 \rowcolor{gray10}
 No documented reactions or intolerances & 0.9167 & 1.0000 & 0.9565 & 11\\
 No reactivity to prescribed drug's ingredients or excipients & 1.0000 & 0.9667 & 0.9831 & 30\\
 \hline
 \textbf{Macro Average} & \textbf{0.9853} & \textbf{0.9893} & \textbf{0.9869} & \textbf{1,000}\\
 \hline
 \end{tabular}
 }
 \label{table:results}
 \end{table} 

Regarding the category classification, the system achieved excellent performance, with an overall macro-averaged precision of 0.9853, recall of 0.9893, and F1 score of 0.9869 (see Table \ref{table:results}). 
Notably, perfect classification (precision = 1.0000, recall = 1.0000, F1 = 1.0000) was achieved for three critical categories: "Direct Active Ingredient Reactivity," "Drug Class Cross-Reactivity with Documented Tolerance," and "Drug Class Cross-Reactivity Without Documented Tolerance." This perfect performance is particularly significant for the cross-reactivity categories, which comprised the majority of cases in the dataset (855 out of 1,000 cases).

The "Chemical-Based Cross-Reactivity to Excipients" category showed excellent performance with a precision of 0.9804, perfect recall of 1.0000, and F1 score of 0.9901. Similar high performance was observed for "Direct Excipient Reactivity" (precision = 1.0000, recall = 0.9583, F1 = 0.9787), with only one case misclassified as "Chemical-Based Cross-Reactivity to Excipients." Importantly, this misclassification occurred between categories with similar non-life-threatening, non-immune-mediated reaction profiles.

For the "No Documented Reactions or Intolerances" category, the system achieved a precision of 0.9167 and perfect recall (1.0000), resulting in an F1 score of 0.9565. Similarly, "No Reactivity to Prescribed Drug's Ingredients or Excipients" showed perfect precision (1.0000) with slightly lower recall (0.9667) and a high F1 score of 0.9831, with one case misclassified as "No Documented Reactions or Intolerances" - both categories representing scenarios with no adverse reactions.

This pattern of errors suggests that while the system occasionally struggles to differentiate between certain similar non-serious reaction types, it maintains robust performance for scenarios with more serious clinical implications.

\begin{table} [h]
 \scriptsize
 \centering
     \caption{Classification Results per Reaction Type (averaged over five runs).} \vspace{1mm}
     \resizebox{1\linewidth}{!}{     \begin{tabular}{p{0.55\linewidth} p{0.1\linewidth} p{0.1\linewidth} p{0.1\linewidth} p{0.15\linewidth}} \rowcolor{black} \textcolor{white}{Reaction Type} & \textcolor{white}{Precision} & \textcolor{white}{Recall} & \textcolor{white}{F1} & \textcolor{white}{Cases} \\
 \hline
 None & 1.0000 & 1.0000 & 1.0000 & 396\\
 \rowcolor{gray10}
 Life-threatening & 1.0000 & 1.0000 & 1.0000 & 133\\ 
 Non life-threatening immune-mediated & 1.0000 & 1.0000 & 1.0000 & 296\\
\rowcolor{gray10}
 Non life-threatening non immune-mediated & 0.9941 & 1.0000 & 1.0000 & 175\\
 \hline
 \textbf{Macro Average} & \textbf{1.0000} & \textbf{1.0000} & \textbf{1.0000} & \textbf{1,000}\\
 \hline
 \end{tabular}
 }
 \label{table:results_react}
 \end{table} 
 
The evaluation of reaction type identification (see Table \ref{table:results_react}) showed perfect accuracy in distinguishing between different severity levels of adverse drug reactions, from life-threatening cases to non-immune-mediated reactions.

\begin{table} [h]
 \scriptsize
 \centering
     \caption{Classification Results per Alert Type (averaged over five runs).} \vspace{1mm}
     \resizebox{1\linewidth}{!}{     \begin{tabular}{p{0.25\linewidth} p{0.15\linewidth}  p{0.15\linewidth} p{0.15\linewidth} p{0.3\linewidth}} \rowcolor{black}  & \textcolor{white}{Ground} & \textcolor{white}{Heliot} &
     \textcolor{white}{Traditional}\\
     \rowcolor{black} \textcolor{white}{Alert Type} & \textcolor{white}{Truth (\%)} & \textcolor{white}{(\%)} & \textcolor{white}{Systems (\%)}\\
 \hline
 No Alert Needed & 396 (39.6\%) & 395 (39.5\%) & 41 (4.1\%)\\
 \rowcolor{gray10}
 Interruptive Alert & 455 (45.5\%) & 455 (45.5\%) & 959 (95.9\%)\\ 
 Non-Interruptive Alert & 149 (14.9\%) & 148 (14.8\%) & 0 (0\%)\\
 \hline
 \end{tabular}
 }
 \label{table:results_alert}
 \end{table} 
 
Regarding alert generation (see Table \ref{table:results_alert}), our approach suggests  improvements over traditional CDSSs, which tend to favor interruptive alerts \cite{hsieh2004characteristics, topaz2015high, topaz2016rising, colicchio2023beyond}. By distinguishing between cases requiring interruptive alerts (39.5\%), those where non-interruptive alerts might suffice (14.8\% vs 14.9\%), and situations where no alert is needed (39.5\% vs 39.6\%), HELIOT shows promise in addressing alert fatigue by potentially reducing interruptive alerts by 50.2\% (395 - 41 No Alert Needed + 148 Non-Interruptive Alert) compared to traditional systems. However, these results need validation in real clinical settings, where alert override rates typically exceed 90\%.

Another key aspect derived from analyzing the results is the system's efficiency. The average execution time was 2.775 seconds per patient to see the overall response on the web application. Importantly, the response is streamed in real-time, similar to how ChatGPT functions, so healthcare professionals can start reading the suggestions immediately as they are generated without significant delays.  
This efficiency stems from several architectural optimizations: TileDB's efficient data retrieval using Zstandard compression \cite{collet2018zstandard} and LRU caching \cite{fegaras2022scalable}, an in-memory synonyms dictionary for quick ingredient resolution, and parallel processing for database queries and translations. These optimizations not only enhance performance but also improve the reliability of the RAG process by maintaining an up-to-date knowledge base and reducing the risk of LLM hallucinations.

Given the obtained results and the discussion above, we can answer our research question as follows:

\begin{prompt}{RQ Answer:}{HELIOT shows promise in providing support during drug administration, offering accurate and efficient identification of potential adverse drug reactions, reducing alert fatigue.  Nonetheless, real-world clinical validation will be essential to confirm its practical effectiveness.}
\end{prompt}

\section{Implications, Limitations and Future Work}
\label{sec:discussion}
This section discusses the practical implications of our findings and outlines limitations and future research directions.

\subsection{Practical Implications and Future Work}
The initial results of HELIOT CDSS suggest advancements in addressing key challenges in adverse drug reaction management, particularly in healthcare settings where clinical information exists primarily in unstructured formats. By leveraging LLMs to interpret clinical narratives and generate contextual alerts, HELIOT demonstrates an approach to reduce alert fatigue while maintaining safety, a significant improvement over traditional rule-based systems that typically generate non-specific alerts for all potentially cross-reactive medications. 

From an implementation perspective, HELIOT's ability to process unstructured clinical notes and generate appropriate alerts makes it particularly valuable across diverse healthcare settings. While some environments have advanced EHR systems that could feed structured data into HELIOT's database, many healthcare facilities worldwide still operate with basic or no EHR infrastructure, relying primarily on unstructured clinical notes. HELIOT's flexible architecture accommodates both scenarios: it can process data from existing EHR systems where available while functioning independently using clinical narratives in settings with limited technological infrastructure. This versatility and its microservices architecture and streaming response mechanism facilitate adaptation to different clinical workflows, from specialized hospitals to primary care practices. Additionally, the system's capability to maintain patient clinical histories could enhance continuity of care across different healthcare settings.

The impact on patient safety and healthcare costs may be significant. By improving the accuracy of adverse drug reaction alerts and reducing alert fatigue, HELIOT could help prevent ADEs while ensuring that critical warnings are not overlooked. All this leads to fewer medication errors, reduced hospital readmissions, and shorter hospital stays. The economic implications of these improvements could be substantial, as medication errors and ADEs impose significant costs on healthcare systems. The resources saved could be redirected to other critical aspects of patient care. 

Looking ahead, several crucial directions for future work emerge. The primary focus will be validating the system with real-world clinical data and scenarios. While our synthetic dataset provided a foundation for initial evaluation, real-world validation is essential to assess the system's effectiveness in handling the complexities of actual clinical narratives, including varied documentation styles and 
medication histories. Another important aspect involves expanding HELIOT's capabilities to handle more complex clinical scenarios, such as multiple drug interactions and patient-specific factors. Such expansion requires enhancing the LLM's ability to interpret nuanced clinical information, such as temporal relationships in medication tolerance and cross-reactivity patterns, often poorly handled by current rule-based systems.

Additionally, future research will focus on optimizing the system's performance and conducting comprehensive user satisfaction studies in various healthcare settings. These evaluations will help understand how HELIOT's approach to alert generation impacts clinical workflow and decision-making in different healthcare environments.

\subsection{Limitations}
\label{sec:threats}
Despite the promising results, HELIOT presents a few limitations. 
The primary limitation of our evaluation is the synthetic nature of the patient dataset used in our experiments. Synthetic data may not fully capture the complexity and variability of real-world patient data. To address this, we involved two healthcare professionals in the ground truth verification process to ensure the synthetic data's accuracy and relevance. 
This approach is supported by recent literature, which highlights how synthetic data can replicate real-world analysis results while maintaining privacy  \cite{kuhnel2024synthetic, giuffre2023harnessing}. Additionally, we plan to continuously update and expand the synthetic dataset, allowing the community to contribute and refine it, thereby improving its realism and utility over time.
Another limitation relates to the HELIOT prototype, specifically to the management of drug leaflets that contain information for multiple pharmaceutical forms. We mitigated this issue by using specific GPT-4o prompts to extract information relevant only to the pharmaceutical form of interest, with healthcare professionals validating the extraction process.
The non-deterministic nature of GPT-4o poses another concern. Response variability can occur, which we mitigated by employing best practices in prompt engineering and running the experiments multiple times to ensure consistency. Specifically, 
we provided clear and specific instructions, using structured templates, and incorporating examples to guide the model's behavior. Then,  
we conducted five separate runs of the experiment, each time recording the precision, recall, and F1-score, to verify that the results were consistent across different runs.

\section{Conclusion}
\label{sec:conclusions}
This paper presented HELIOT CDSS, a novel approach to adverse drug reaction management that leverages LLMs to interpret unstructured clinical narratives. While traditional CDSSs often struggle with free-text clinical notes and generate excessive alerts due to their rule-based nature, HELIOT demonstrated promise  in processing natural language descriptions of adverse reactions and generating more contextual alerts. Our initial evaluation showed encouraging results in a controlled setting, particularly in the system's ability to process clinical narratives and distinguish between cases requiring different types of alerts. 
Integrating comprehensive pharmaceutical data repositories and using LLMs for text interpretation represents a step forward in handling the complexity of adverse drug reaction information, often in unstructured formats across different healthcare settings. 

The potential impact of this approach could be considerable, particularly in healthcare environments with varying levels of EHR integration. By improving the interpretation of clinical narratives and providing more contextual alerts, HELIOT could help reduce alert fatigue while maintaining patient safety. All this could lead to better clinical outcomes and cost savings by preventing adverse drug events and reducing unnecessary alert overrides. 

Future work will validate the system with real-world clinical data and scenarios, particularly testing its performance with actual clinical narratives and 
medication histories. Additionally, efforts will be made to enhance the system's capabilities in handling more complex clinical scenarios and to evaluate its impact on clinical workflow in various healthcare settings.

\bibliographystyle{unsrt}
\bibliography{bibliography}

\begin{thebibliography}{10}

\bibitem{elliott2021economic}
Rachel~Ann Elliott, Elizabeth Camacho, Dina Jankovic, Mark~J Sculpher, and Rita Faria.
\newblock Economic analysis of the prevalence and clinical and economic burden of medication error in england.
\newblock {\em BMJ Quality \& Safety}, 30(2):96--105, 2021.

\bibitem{ahsani2022interventions}
Ehsan Ahsani-Estahbanati, Vladimir Sergeevich~Gordeev, and Leila Doshmangir.
\newblock Interventions to reduce the incidence of medical error and its financial burden in health care systems: A systematic review of systematic reviews.
\newblock {\em Frontiers in medicine}, 9:875426, 2022.

\bibitem{sutton2020overview}
Reed~T Sutton, David Pincock, Daniel~C Baumgart, Daniel~C Sadowski, Richard~N Fedorak, and Karen~I Kroeker.
\newblock An overview of clinical decision support systems: benefits, risks, and strategies for success.
\newblock {\em NPJ digital medicine}, 3(1):17, 2020.

\bibitem{kwan2020computerised}
Janice~L Kwan, Lisha Lo, Jacob Ferguson, Hanna Goldberg, Juan~Pablo Diaz-Martinez, George Tomlinson, Jeremy~M Grimshaw, and Kaveh~G Shojania.
\newblock Computerised clinical decision support systems and absolute improvements in care: meta-analysis of controlled clinical trials.
\newblock {\em Bmj}, 370, 2020.

\bibitem{sarkar2020effective}
Urmimala Sarkar and Lipika Samal.
\newblock How effective are clinical decision support systems?, 2020.

\bibitem{meunier2023barriers}
Pierre-Yves Meunier, Camille Raynaud, Emmanuelle Guimaraes, Fran{\c{c}}ois Gueyffier, and Laurent Letrilliart.
\newblock Barriers and facilitators to the use of clinical decision support systems in primary care: a mixed-methods systematic review.
\newblock {\em The Annals of Family Medicine}, 21(1):57--69, 2023.

\bibitem{quan2023usefulness}
Paola~Leonor Quan, Sergio S{\'a}nchez-Fern{\'a}ndez, Luc{\'\i}a Parrado~Gil, Alfonso Calvo~Alonso, Jos{\'e}~Miguel Bodero~S{\'a}nchez, Ana Ortega~Eslava, Marta Luri, and Gabriel Gastaminza~Lasarte.
\newblock Usefulness of drug allergy alert systems: Present and future.
\newblock {\em Current Treatment Options in Allergy}, 10(4):413--427, 2023.

\bibitem{colicchio2023beyond}
Tiago~K Colicchio and James~J Cimino.
\newblock Beyond the override: Using evidence of previous drug tolerance to suppress drug allergy alerts; a retrospective study of opioid alerts.
\newblock {\em Journal of Biomedical Informatics}, 147:104508, 2023.

\bibitem{van2021optimizing}
Bethany~A Van~Dort, Wu~Yi Zheng, Vivek Sundar, and Melissa~T Baysari.
\newblock Optimizing clinical decision support alerts in electronic medical records: a systematic review of reported strategies adopted by hospitals.
\newblock {\em Journal of the American Medical Informatics Association}, 28(1):177--183, 2021.

\bibitem{westerbeek2021barriers}
Leonie Westerbeek, Kimberley~J Ploegmakers, Gert-Jan De~Bruijn, Annemiek~J Linn, Julia~CM van Weert, Joost~G Daams, Nathalie van~der Velde, Henk~C van Weert, Ameen Abu-Hanna, and Stephanie Medlock.
\newblock Barriers and facilitators influencing medication-related cdss acceptance according to clinicians: a systematic review.
\newblock {\em International Journal of Medical Informatics}, 152:104506, 2021.

\bibitem{van2022overall}
Greet Van De~Sijpe, Charlotte Quintens, Karolien Walgraeve, Eva Van~Laer, Jens Penny, Greet De~Vlieger, Rik Schrijvers, Paul De~Munter, Veerle Foulon, Minne Casteels, et~al.
\newblock Overall performance of a drug--drug interaction clinical decision support system: quantitative evaluation and end-user survey.
\newblock {\em BMC Medical Informatics and Decision Making}, 22(1):48, 2022.

\bibitem{topaz2016rising}
Maxim Topaz, Diane~L Seger, Sarah~P Slight, Foster Goss, Kenneth Lai, Paige~G Wickner, Kimberly Blumenthal, Neil Dhopeshwarkar, Frank Chang, David~W Bates, et~al.
\newblock Rising drug allergy alert overrides in electronic health records: an observational retrospective study of a decade of experience.
\newblock {\em Journal of the American Medical Informatics Association}, 23(3):601--608, 2016.

\bibitem{topaz2015high}
Maxim Topaz, Diane~L Seger, Kenneth Lai, Paige~G Wickner, Foster Goss, Neil Dhopeshwarkar, Frank Chang, David~W Bates, and Li~Zhou.
\newblock High override rate for opioid drug-allergy interaction alerts: current trends and recommendations for future.
\newblock In {\em MEDINFO 2015: eHealth-enabled Health}, pages 242--246. IOS Press, 2015.

\bibitem{hsieh2004characteristics}
Tyken~C Hsieh, Gilad~J Kuperman, Tonushree Jaggi, Patricia Hojnowski-Diaz, Julie Fiskio, Deborah~H Williams, David~W Bates, and Tejal~K Gandhi.
\newblock Characteristics and consequences of drug allergy alert overrides in a computerized physician order entry system.
\newblock {\em Journal of the American Medical Informatics Association}, 11(6):482--491, 2004.

\bibitem{achiam2023gpt}
Josh Achiam, Steven Adler, Sandhini Agarwal, Lama Ahmad, Ilge Akkaya, Florencia~Leoni Aleman, Diogo Almeida, Janko Altenschmidt, Sam Altman, Shyamal Anadkat, et~al.
\newblock Gpt-4 technical report.
\newblock {\em arXiv preprint arXiv:2303.08774}, 2023.

\bibitem{tripathi2024efficient}
Satvik Tripathi, Rithvik Sukumaran, and Tessa~S Cook.
\newblock Efficient healthcare with large language models: optimizing clinical workflow and enhancing patient care.
\newblock {\em Journal of the American Medical Informatics Association}, 31(6):1436--1440, 2024.

\bibitem{rios2024evaluation}
Alejandro R{\'\i}os-Hoyo, Naing~Lin Shan, Anran Li, Alexander~T Pearson, Lajos Pusztai, and Frederick~M Howard.
\newblock Evaluation of large language models as a diagnostic aid for complex medical cases.
\newblock {\em Frontiers in Medicine}, 11:1380148, 2024.

\bibitem{de2025llms}
Gabriele De~Vito, Filomena Ferrucci, and Athanasios Angelakis.
\newblock Llms for drug-drug interaction prediction: A comprehensive comparison.
\newblock {\em arXiv preprint arXiv:2502.06890}, 2025.

\bibitem{de2024assessing}
Gabriele De~Vito.
\newblock Assessing healthcare software built using iot and llm technologies.
\newblock In {\em Proceedings of the 28th International Conference on Evaluation and Assessment in Software Engineering}, pages 476--481, 2024.

\bibitem{roy2022drug}
Romy Roy, Shamsudheen Marakkar, Munawar~P Vayalil, Alisha Shahanaz, Athira~Panicker Anil, Shameer Kunnathpeedikayil, Ishaan Rawal, Kavya Shetty, Zahrah Shameer, Saraswathi Sathees, et~al.
\newblock Drug-food interactions in the era of molecular big data, machine intelligence, and personalized health.
\newblock {\em Recent Advances in Food Nutrition \& Agriculture}, 13(1):27--50, 2022.

\bibitem{corny2020machine}
Jennifer Corny, Asok Rajkumar, Olivier Martin, Xavier Dode, Jean-Patrick Lajonch{\`e}re, Olivier Billuart, Yvonnick B{\'e}zie, and Anne Buronfosse.
\newblock A machine learning--based clinical decision support system to identify prescriptions with a high risk of medication error.
\newblock {\em Journal of the American Medical Informatics Association}, 27(11):1688--1694, 2020.

\bibitem{luri2022systematic}
Marta Luri, Leire Leache, Gabriel Gastaminza, Antonio Idoate, and Ana Ortega.
\newblock A systematic review of drug allergy alert systems.
\newblock {\em International journal of medical informatics}, 159:104673, 2022.

\bibitem{calvo2022ontopharma}
Elena Calvo-Cidoncha, Concepci{\'o}n Camacho-Hernando, Faust Feu, Xavier Pastor-Duran, Carles Codina-Jan{\'e}, and Raimundo Lozano-Rub{\'\i}.
\newblock Ontopharma: ontology based clinical decision support system to reduce medication prescribing errors.
\newblock {\em BMC Medical Informatics and Decision Making}, 22(1):238, 2022.

\bibitem{rozenblum2020using}
Ronen Rozenblum, Rosa Rodriguez-Monguio, Lynn~A Volk, Katherine~J Forsythe, Sara Myers, Maria McGurrin, Deborah~H Williams, David~W Bates, Gordon Schiff, and Enrique Seoane-Vazquez.
\newblock Using a machine learning system to identify and prevent medication prescribing errors: a clinical and cost analysis evaluation.
\newblock {\em The Joint Commission Journal on Quality and Patient Safety}, 46(1):3--10, 2020.

\bibitem{vaswani_2017}
Ashish Vaswani, Noam Shazeer, Niki Parmar, Jakob Uszkoreit, Llion Jones, Aidan~N Gomez, {\L}ukasz Kaiser, and Illia Polosukhin.
\newblock Attention is all you need.
\newblock volume~30, 2017.

\bibitem{gpt3_2020}
Tom Brown, Benjamin Mann, Nick Ryder, Melanie Subbiah, Jared~D Kaplan, Prafulla Dhariwal, Arvind Neelakantan, Pranav Shyam, Girish Sastry, Amanda Askell, et~al.
\newblock Language models are few-shot learners.
\newblock volume~33, pages 1877--1901, 2020.

\bibitem{bert_2018}
Jacob Devlin, Ming-Wei Chang, Kenton Lee, and Kristina Toutanova.
\newblock Bert: Pre-training of deep bidirectional transformers for language understanding.
\newblock In {\em Proceedings of the 2019 conference of the North American chapter of the association for computational linguistics: human language technologies, volume 1 (long and short papers)}, pages 4171--4186, 2019.

\bibitem{chung2022scaling}
Hyung~Won Chung, Le~Hou, Shayne Longpre, Barret Zoph, Yi~Tay, William Fedus, Yunxuan Li, Xuezhi Wang, Mostafa Dehghani, Siddhartha Brahma, et~al.
\newblock Scaling instruction-finetuned language models.
\newblock {\em Journal of Machine Learning Research}, 25(70):1--53, 2024.

\bibitem{touvron2023llama}
Hugo Touvron, Thibaut Lavril, Gautier Izacard, Xavier Martinet, Marie-Anne Lachaux, Timoth{\'e}e Lacroix, Baptiste Rozi{\`e}re, Naman Goyal, Eric Hambro, Faisal Azhar, et~al.
\newblock Llama: Open and efficient foundation language models, 2023.

\bibitem{scao_2022_bloom}
BigScience Workshop, Teven~Le Scao, Angela Fan, Christopher Akiki, Ellie Pavlick, Suzana Ili{\'c}, Daniel Hesslow, Roman Castagn{\'e}, Alexandra~Sasha Luccioni, Fran{\c{c}}ois Yvon, et~al.
\newblock Bloom: A 176b-parameter open-access multilingual language model, 2022.

\bibitem{zeng_2022_glm}
Aohan Zeng, Xiao Liu, Zhengxiao Du, Zihan Wang, Hanyu Lai, Ming Ding, Zhuoyi Yang, Yifan Xu, Wendi Zheng, Xiao Xia, et~al.
\newblock Glm-130b: An open bilingual pre-trained model, 2022.

\bibitem{chen2021evaluating}
Mark Chen, Jerry Tworek, Heewoo Jun, Qiming Yuan, Henrique Ponde De~Oliveira Pinto, Jared Kaplan, Harri Edwards, Yuri Burda, Nicholas Joseph, Greg Brockman, et~al.
\newblock Evaluating large language models trained on code, 2021.

\bibitem{KASNECI2023102274}
Enkelejda Kasneci, Kathrin Se{\ss}ler, Stefan K{\"u}chemann, Maria Bannert, Daryna Dementieva, Frank Fischer, Urs Gasser, Georg Groh, Stephan G{\"u}nnemann, Eyke H{\"u}llermeier, et~al.
\newblock Chatgpt for good? on opportunities and challenges of large language models for education.
\newblock {\em Learning and individual differences}, 103:102274, 2023.

\bibitem{zhao2023survey}
Wayne~Xin Zhao, Kun Zhou, Junyi Li, Tianyi Tang, Xiaolei Wang, Yupeng Hou, Yingqian Min, Beichen Zhang, Junjie Zhang, Zican Dong, et~al.
\newblock A survey of large language models, 2023.

\bibitem{shanahan2023talking}
Murray Shanahan.
\newblock Talking about large language models, 2023.

\bibitem{white2023prompt}
Jules White, Quchen Fu, Sam Hays, Michael Sandborn, Carlos Olea, Henry Gilbert, Ashraf Elnashar, Jesse Spencer-Smith, and Douglas~C Schmidt.
\newblock A prompt pattern catalog to enhance prompt engineering with chatgpt.
\newblock {\em arXiv preprint arXiv:2302.11382}, 2023.

\bibitem{wendler2024llamas}
Chris Wendler, Veniamin Veselovsky, Giovanni Monea, and Robert West.
\newblock Do llamas work in english? on the latent language of multilingual transformers.
\newblock In {\em Proceedings of the 62nd Annual Meeting of the Association for Computational Linguistics (Volume 1: Long Papers)}, pages 15366--15394, 2024.

\bibitem{papadopoulos2016tiledb}
Stavros Papadopoulos, Kushal Datta, Samuel Madden, and Timothy Mattson.
\newblock The tiledb array data storage manager.
\newblock {\em Proceedings of the VLDB Endowment}, 10(4):349--360, 2016.

\bibitem{devito_github}
G.~De~Vito, A.~Angelakis, and F.~Ferrucci.
\newblock Online repository, 2024.

\bibitem{flahault2005sample}
Antoine Flahault, Michel Cadilhac, and Guy Thomas.
\newblock Sample size calculation should be performed for design accuracy in diagnostic test studies.
\newblock {\em Journal of clinical epidemiology}, 58(8):859--862, 2005.

\bibitem{cohen1960coefficient}
Jacob Cohen.
\newblock A coefficient of agreement for nominal scales.
\newblock {\em Educational and psychological measurement}, 20(1):37--46, 1960.

\bibitem{collet2018zstandard}
Yann Collet and Murray Kucherawy.
\newblock Zstandard compression and the application/zstd media type.
\newblock Technical report, 2018.

\bibitem{kuhnel2024synthetic}
Lisa K{\"u}hnel, Julian Schneider, Ines Perrar, Tim Adams, Sobhan Moazemi, Fabian Prasser, Ute N{\"o}thlings, Holger Fr{\"o}hlich, and Juliane Fluck.
\newblock Synthetic data generation for a longitudinal cohort study--evaluation, method extension and reproduction of published data analysis results.
\newblock {\em Scientific Reports}, 14(1):14412, 2024.

\bibitem{giuffre2023harnessing}
Mauro Giuffr{\`e} and Dennis~L Shung.
\newblock Harnessing the power of synthetic data in healthcare: innovation, application, and privacy.
\newblock {\em NPJ digital medicine}, 6(1):186, 2023.

\bibitem{lee2022towards}
Min~Hun Lee, Daniel~P Siewiorek, Asim Smailagic, Alexandre Bernardino, and Sergi Berm{\'u}dez~i Badia.
\newblock Towards efficient annotations for a human-ai collaborative, clinical decision support system: A case study on physical stroke rehabilitation assessment.
\newblock In {\em Proceedings of the 27th International Conference on Intelligent User Interfaces}, pages 4--14, 2022.

\bibitem{pais2024large}
Cristobal Pais, Jianfeng Liu, Robert Voigt, Vin Gupta, Elizabeth Wade, and Mohsen Bayati.
\newblock Large language models for preventing medication direction errors in online pharmacies.
\newblock {\em Nature Medicine}, pages 1--9, 2024.

\bibitem{valente2022interpretability}
Francisco Valente, Sim{\~a}o Paredes, Jorge Henriques, Teresa Rocha, Paulo de~Carvalho, and Jo{\~a}o Morais.
\newblock Interpretability, personalization and reliability of a machine learning based clinical decision support system.
\newblock {\em Data Mining and Knowledge Discovery}, 36(3):1140--1173, 2022.

\bibitem{fleiss2013statistical}
Joseph~L Fleiss, Bruce Levin, and Myunghee~Cho Paik.
\newblock {\em Statistical methods for rates and proportions}.
\newblock john wiley \& sons, 2013.

\bibitem{zampetti2022empirical}
Fiorella Zampetti, Ritu Kapur, Massimiliano Di~Penta, and Sebastiano Panichella.
\newblock An empirical characterization of software bugs in open-source cyber--physical systems.
\newblock {\em Journal of Systems and Software}, 192:111425, 2022.

\bibitem{fegaras2022scalable}
Leonidas Fegaras, Tanvir~Ahmed Khan, Md~Hasanuzzaman Noor, and Tanzima Sultana.
\newblock Scalable tensors for big data analytics.
\newblock In {\em 2022 IEEE International Conference on Big Data (Big Data)}, pages 107--114. IEEE, 2022.

\end{thebibliography}

\end{document}